\documentclass{article}

\usepackage[preprint]{corl_2026} 

\usepackage{graphicx}
\usepackage{wrapfig}
\usepackage{amsmath,amssymb}
\usepackage{booktabs}
\usepackage{multirow}
\usepackage{enumitem}

\newcommand{\secv}{\vspace{-0.6em}}

\newcommand{\myparagraph}[1]{\vspace{-0.1em}\noindent\textbf{#1}}
\newcommand{\ms}[2]{#1 {\small $\!\pm\!$ #2}}

\title{A Minimalist Retargeting-Guided Reinforcement Learning Recipe for Dexterous Manipulation}

\author{
  Yunhai Feng$^1$\enspace
  Natalie Leung$^1$\enspace
  Jiaxuan Wang$^1$\enspace
  Lujie Yang$^2$\enspace
  Haozhi Qi$^2$\enspace
  Preston Culbertson$^1$
  \\[0.2em]
  $^1$Cornell University\quad
  $^2$Amazon FAR
}

\begin{document}
\maketitle

\begin{figure}[h]
    \centering
    \vspace{-3em}
    \includegraphics[width=\textwidth]{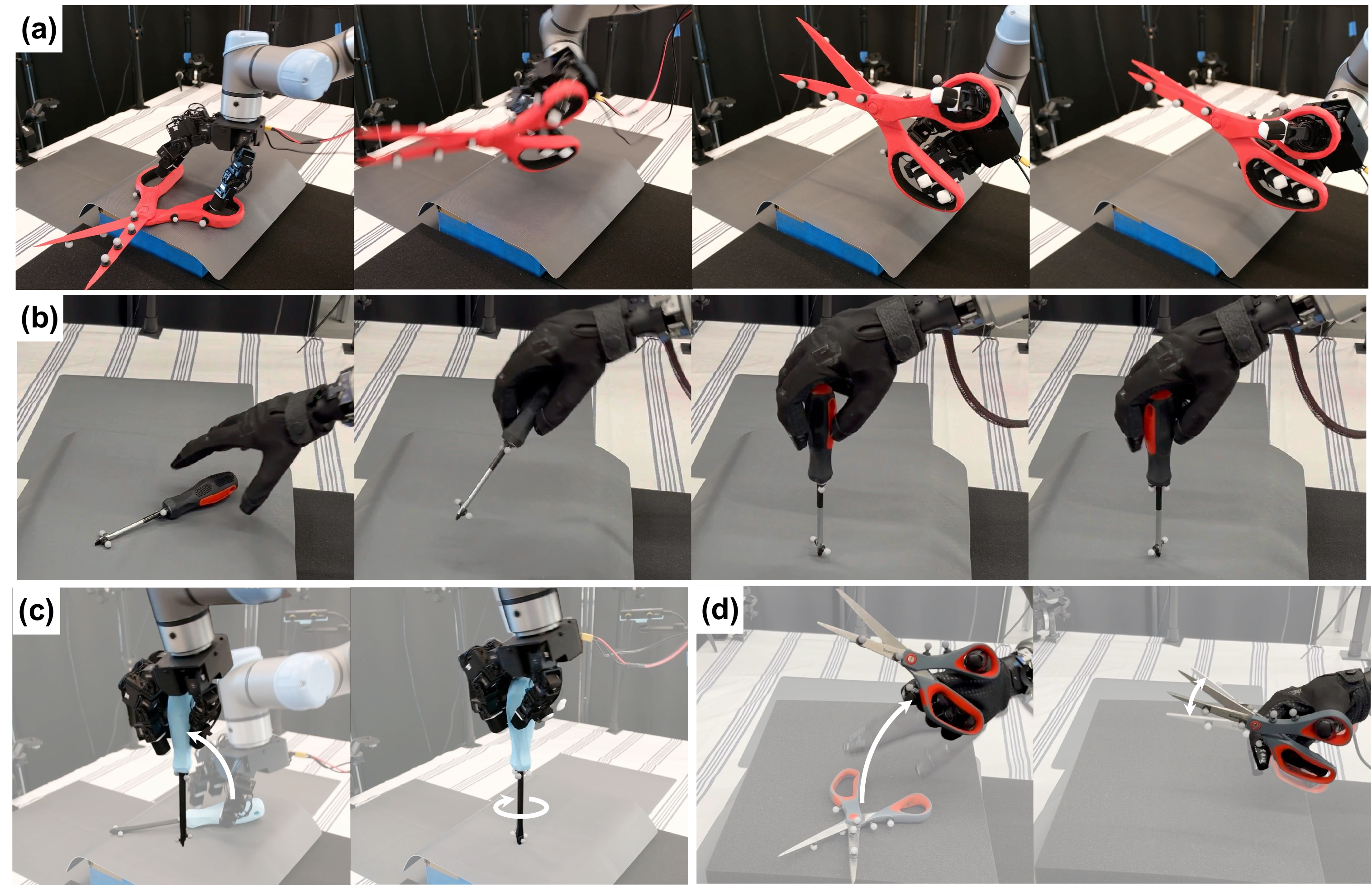}
    \vspace{-1em}
    \caption{\textbf{Dexterous Tool Use Tasks.} We present a minimalist retargeting-guided RL pipeline for learning dexterous manipulation from a single human demonstration. We evaluate our method on four contact-rich task-hand settings across two robot hands: (a) \texttt{LEAP-Scissors}, (b) \texttt{WUJI-Screwdriver}, (c) \texttt{LEAP-Screwdriver}, and (d) \texttt{WUJI-Scissors}.}
    \label{fig:teaser}
\end{figure}

\vspace{-0.45em}
\begin{abstract}
  Recent work in humanoid whole-body control has found success with a simple recipe: retarget human motion to robot kinematic references, then train policies via reinforcement learning (RL) to track them. But how does this recipe transfer to dexterous manipulation?
  The answer is not obvious, as manipulation involves complex, contact-rich dynamics and requires delicate regulation of contact modes and forces.
  We present \textsc{Regrind}, a minimalist retargeting-guided RL pipeline that learns dexterous manipulation policies from a single human demonstration. \textsc{Regrind} retargets human hand-object motion to a robot reference that preserves hand-object spatial and contact relationships, trains a residual RL policy in simulation to track object-centric keypoints along that reference, and transfers the resulting policy zero-shot to hardware with careful system identification. The resulting policies produce fluid, human-like behavior on two different multi-fingered hands across contact-rich tool-use tasks, including operating a pair of scissors and turning a screwdriver. Through systematic hardware experiments, we identify and analyze the key factors that govern sim-to-real transfer in dexterous manipulation, offering practical guidance for retargeting-based learning in contact-rich settings. 
  Videos and code are available at \url{https://yunhaifeng.com/REGRIND}.
\end{abstract}



\secv
\section{Introduction}
\secv
Humanoid robots and anthropomorphic hands share the human body's kinematic structure, opening the possibility of learning skills directly from the vast amount of human motion data. In humanoid whole-body control, this has led to a simple and increasingly effective recipe: retarget a human motion reference to the robot, train a policy in simulation to track the retargeted reference, and deploy the policy on hardware~\citep{liao2025beyondmimic,yang2025omniretarget}. Dexterous manipulation, however, has largely followed a different path. Current manipulation skills are often learned from teleoperated demonstrations collected directly on real robots~\citep{zhao2023learning,wang2024dexcap,cheng2025open}, with limited use of either human motion data or massive simulation.

While recent work has explored using humanoid-style retargeting pipelines for dexterous hands, results have largely been confined to simulation or open-loop replays of simulation motions~\citep{li2025maniptrans, mandi2025dexmachina}. One hypothesis for this limited real-world success is that simple kinematic retargeting, while closely matching the human hand pose, ignores the hand-object interaction and produces physically implausible trajectories that are a poor reference for downstream RL. This raises a natural question: \textit{does preserving interactions during retargeting improve RL for contact-rich dexterous manipulation?}

In this work, we answer this question by presenting \textsc{Regrind}: \textbf{RE}targeting-\textbf{G}uided \textbf{R}e\textbf{IN}forcement learning for \textbf{D}exterous manipulation, a minimalist pipeline for learning contact-rich dexterous manipulation skills from a single human demonstration. We first retarget human hand-object motion to an interaction-preserving robot reference that maintains the spatial relationships between the hand and the object. To mitigate the scarcity of multi-fingered dexterous manipulation data, we dynamically augment the retargeted reference with perturbed initial configurations to provide broader coverage of the task space. We then train a residual RL policy in simulation to track object-centric keypoints along this reference, using the retargeted trajectories as both a motion prior and reset distribution for exploration.
The resulting policies transfer to real multi-fingered hands on challenging tool-use tasks, including operating a pair of scissors and turning a screwdriver, which require coordinated finger motion, complex contact transitions, and interactions that leverage friction.

At the same time, our study reveals an important distinction between locomotion and manipulation. Although retargeting-based learning can produce highly capable manipulation policies~\citep{li2025maniptrans,mandi2025dexmachina}, dexterous manipulation exhibits substantially greater sensitivity to sim-to-real discrepancies. Manipulation hinges on rich, sustained contact between the robot and the object, and small modeling errors in friction, compliance, or geometry compound quickly when the policy's success depends on those interactions. Our results suggest that successful sim-to-real transfer requires not only interaction-preserving retargeting but also thoughtful design of observation and action spaces, appropriate domain randomization and curricula, and careful system identification. Together, these ingredients form a practical recipe for sim-to-real dexterous manipulation.

Our contributions are as follows:
\begin{itemize}[itemsep=1pt, topsep=0pt, leftmargin=15pt, parsep=0pt]
    \item We propose a general and minimalist framework for learning dexterous manipulation from a single human demonstration. We release the code and data for the community to reproduce the results and build upon our work.
    \item We apply a simple yet effective data augmentation strategy that exposes the RL policy to diverse trajectories for improved robustness and generalization. 
    \item We evaluate our method on challenging contact-rich dexterous manipulation tasks and demonstrate its superior performance over existing methods, especially the value of interaction-preserving motion retargeting.
\end{itemize}

\secv
\section{Related Work}
\secv
\myparagraph{Dexterous Manipulation with Reinforcement Learning.} Reinforcement learning (RL) has driven much of the progress in dexterous manipulation. While some works have trained manipulation policies directly in the real world~\citep{zhu2019dexterous, nagabandi2020deep, luo2025precise, dodeja2026life}, the high cost of real-world interaction and the sample inefficiency of RL have made training in simulation followed by sim-to-real transfer a more practical approach. Such sim-to-real efforts have progressed from reorienting a cube and solving a Rubik's cube~\citep{andrychowicz2020learning, handa2023dextreme, akkaya2019solving} to reorienting diverse objects~\citep{chen2023visual,qi2023hand,qi2023general,wang2025lessons}, and, increasingly, to contact-rich, bimanual, and tool-use tasks~\citep{lin2024twisting, lin2025sim, kedia2026simtoolreal}. However, scaling RL across tasks remains bottlenecked by reward engineering and exploration. A growing line of work injects external priors such as retargeted reference trajectories~\citep{lum2025crossing, mandi2025dexmachina} or coarse plans from vision-language-model~\citep{de2025scaffolding} to improve learning efficiency. We follow this direction, using human hand motion---faithfully transferred via interaction-preserving retargeting---as a prior that simplifies reward design and biases exploration toward reliable human-like strategies.

\myparagraph{Dexterous Manipulation with Imitation Learning.}
Propelled by scalable data collection systems~\citep{zhao2023learning, cheng2025open, wang2024dexcap} and expressive policy classes~\citep{chi2025diffusion, black2024pi0}, imitation learning (IL) has become the central paradigm for real-world robot manipulation by learning visuomotor policies directly from real-robot demonstrations and thereby sidestepping the sim-to-real gap. However, IL relies on accurate on-robot action data, which is difficult to capture for high-degree-of-freedom dexterous hands. A range of data collection pipelines have emerged for multi-fingered hands, including systems that capture human hand motion with motion capture gloves, VR headsets, or RGB cameras and then retarget it onto robot hands~\citep{yin2025dexteritygen,wang2024dexcap,shaw2024bimanual,qin2023anyteleop}, wearable hand exoskeletons that mechanically couple human and robot fingers for portable, in-the-wild collection~\citep{fang2025dexop, xu2025dexumi, tao2025dexwild}, and methods that augment human demonstrations with physics-based simulation~\citep{jiang2025dexmimicgen}. Our pipeline does not require collecting robot-specific data; instead, it transfers raw human motion to robot trajectories via motion retargeting, placing it close to work that leverages motion-capture and video data as broad sources of human motion~\citep{mandi2025dexmachina,li2025maniptrans,qin2022dexmv,zheng2026egoscale,zhang2026unidex}.

\myparagraph{Motion Retargeting from Human Data.}
Retargeting human motion onto the robot and training policies in simulation to track the reference has become an increasingly common recipe for humanoid whole-body control~\citep{he2024omnih2o,he2025asap,liao2025beyondmimic,yang2025omniretarget}.
For dexterous hands, retargeting has mainly supported teleoperation and demonstration generation through per-frame kinematic optimization~\citep{handa2020dexpilot,qin2023anyteleop} or learned mappings from human video~\citep{sivakumar2022robotic,shaw2023videodex}.
More recent pipelines incorporate object context through functional retargeting to demonstrated object states~\citep{mandi2025dexmachina}, physics-informed retargeting~\citep{pan2025spider}, learned human-to-robot transfer~\citep{park2025learning,li2025maniptrans}, trajectory-guided refinement in simulation~\citep{chen2025vividex}, or sparse object-centric rewards and hand poses to guide RL~\citep{lum2025crossing}.
However, hand-centric retargeting can still yield physically implausible contact 
structures, and prior object-aware methods either target dataset generation and 
imitation, emphasize simulation-only tracking, or rely on sparse object or hand 
cues rather than dense interaction structure.
We adapt the interaction mesh formulation~\citep{ho2010spatial,yang2025omniretarget} from humanoid loco-manipulation to dexterous hand--object manipulation and use the interaction-preserving reference as an exploratory restart distribution for efficient closed-loop sim-to-real RL.

\secv
\section{Method: Retargeting-Guided Reinforcement Learning}
\secv
\begin{figure*}[t!]
    \centering
    \vspace{-2em}
    \includegraphics[width=0.99\textwidth]{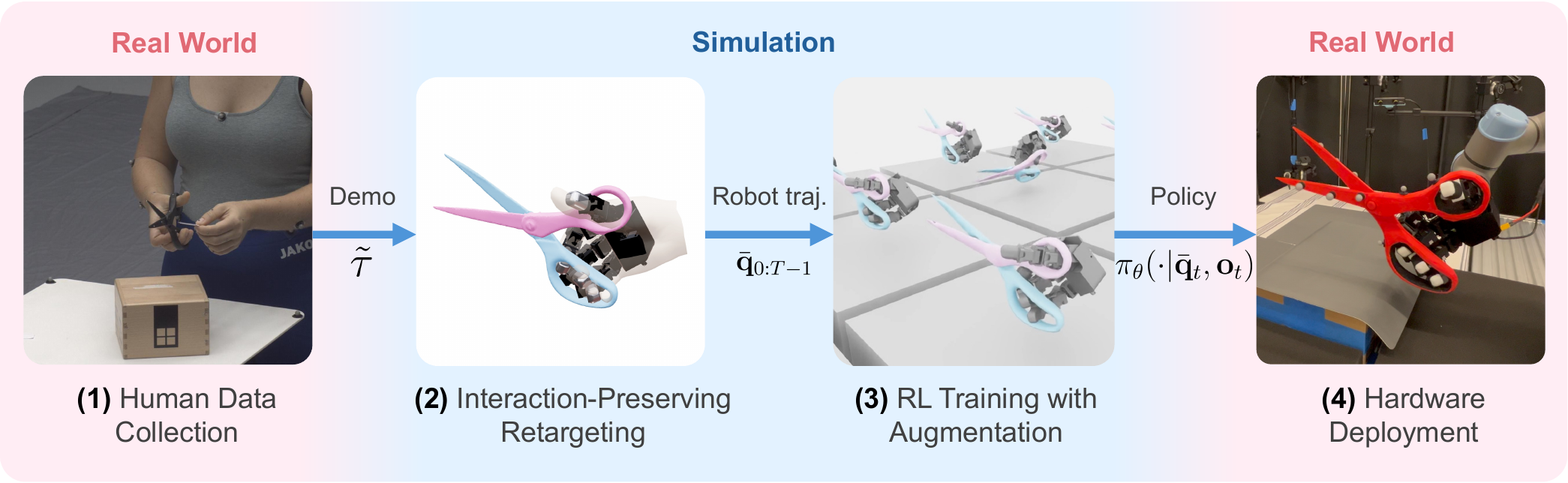}
    \caption{\textbf{Method Overview.} We follow a real-to-sim-to-real pipeline to learn an RL policy from 3D human data. (1) Given a human demonstration with hand and object poses, we (2) retarget it to the robot with interaction-aware motion retargeting, (3) train a policy to track the retargeted trajectory with large-scale RL in simulation, and then (4) transfer the policy zero-shot to the real world. We dynamically augment the reference trajectory during RL training to make the policy generalizable to different initial configurations.
    }
    \vspace{-1em} \label{fig:method_overview}
\end{figure*}

\subsection{Problem Statement}
\label{sec:problem_statement}

Robot policy learning with reinforcement learning (RL) can be formulated as a finite-horizon Markov Decision Process (MDP) defined by $\langle\mathcal{S}, \mathcal{A}, P, \rho_0, r, T\rangle$, where $\mathcal{S}$ and $\mathcal{A}$ are continuous state and action spaces, $P:\mathcal{S}\times\mathcal{A}\rightarrow\Delta(\mathcal{S})$ is the transition dynamics, $\rho_0$ is the initial state distribution, $r:\mathcal{S}\times\mathcal{A}\rightarrow[0, 1]$ is the reward function, and $T$ is the horizon. The goal is to find the optimal policy $\pi^*:\mathcal{S}\rightarrow\mathcal{A}$ that maximizes the expected cumulative reward:
\begin{equation}
    \label{eq:rl}
    \pi^* = {\arg\max}_{\pi}\, \mathbb{E}_{\tau\sim\pi, s_0\sim\rho_0} \left[ \sum_{t=0}^{T-1}\gamma^t r(s_t, a_t) \right],
\end{equation}
where $\tau = (s_0, a_0, s_1, a_1, \ldots, s_{T-1}, a_{T-1})$ is a trajectory sampled from the policy $\pi$, and $\gamma\in[0, 1)$ is the discount factor.

The manipulation task is specified by a human demonstration $\tilde{\tau}=(\tilde{s}_0, \tilde{s}_1, \ldots, \tilde{s}_{T-1})$, and the reward encourages the robot to track the human demonstration, e.g., $r(s_t, a_t) = \exp\left(-\mathrm{dist}(s_t, \tilde{s}_t)\right)$ with some distance metric $\mathrm{dist}(\cdot, \cdot)$. Here, $\tilde{s}_t\in\tilde{\mathcal{S}}$ is the demonstration state that contains information about the object and the human hand at timestep $t$. Note that $\mathcal{S}$ and $\tilde{\mathcal{S}}$ are two different spaces because of the embodiment gap between the robot and the human. For instance, $s_t\in\mathcal{S}$ contains robot joint positions while $\tilde{s}_t\in\tilde{\mathcal{S}}$ may only provide the hand keypoints. We assume the object state representation is the same in both spaces, thus, a reasonable distance metric can be the distance between two object poses in $\mathrm{SE}(3)$.

One challenge of RL is exploration. \citet{kakade2002approximately} introduced the concept of exploratory restart distribution to guide the exploration process, which is later popularized by DeepMimic~\citep{peng2018deepmimic} as Reference State Initialization (RSI).
Instead of optimizing the objective in Eq.~\eqref{eq:rl}, the policy is optimized with respect to an exploratory restart distribution $\mu$:
\begin{equation}
    \label{eq:rl_restart}
    \max_{\pi}\, \mathbb{E}_{\tau\sim\pi, \textcolor{red}{s_0\sim\mu}} \left[ \sum_{t=0}^{T-1}\gamma^t r(s_t, a_t) \right].
\end{equation}

\begin{wrapfigure}[25]{r}{0.4\textwidth}
    \centering
    \includegraphics[width=\linewidth]{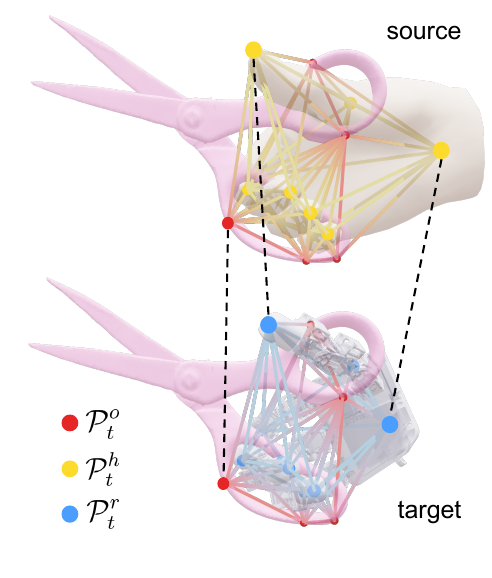}
    \caption{\textbf{Interaction meshes.} We construct the source interaction mesh with the object and hand keypoints in the demonstration (top), and the target interaction mesh with the same object keypoints and the robot keypoints that correspond to the hand keypoints (bottom). The dashed lines connect the corresponding keypoints in the two meshes.
    }
    \label{fig:interaction_mesh}
\end{wrapfigure}

It was also shown that it is important for $\mu$ to be close to the state visitation distribution $d_{\rho_0}^{\pi^*}$ of the optimal policy $\pi^*$\citep{kakade2002approximately}, motivating a natural choice of $\mu$ as the distribution that supports the states from the human demonstration in our case. Due to the embodiment gap discussed earlier, we convert the demonstration states $\tilde{s}_t$ to the robot states $s_t$ using a motion retargeting process and use the retargeted trajectory as the exploratory restart distribution.

The rest of the section is organized as follows. We first introduce the interaction-aware motion retargeting process to convert the human demonstration to a robot trajectory which preserves the hand-object interaction semantics in Section \ref{sec:motion_retargeting}. Then, we introduce the RL training pipeline with the retargeted robot trajectory as the reference motion in Section \ref{sec:policy_learning}.
The full pipeline of our method is illustrated in Fig.~\ref{fig:method_overview}. 

\subsection{Interaction-Aware Motion Retargeting}
\label{sec:motion_retargeting}

The goal of motion retargeting is to convert the human demonstration into a robot trajectory that mimics the human motion. The demonstration consists of the MANO hand keypoints~\cite{romero2022mano} and the object configuration at each timestep, where the object configuration is specified by a 6D pose (and additional joint position(s) if the object is articulated).

We formulate the problem of motion retargeting as an optimization problem on the robot configuration $\mathbf{q}_{0:T-1}=\{(\mathbf{T}_t^r, \mathbf{q}_t^r)\}_{t=0}^{T-1}$, where $\mathbf{T}_t^r$ is the robot wrist pose and $\mathbf{q}_t^r$ is the vector of robot joint positions at timestep $t$.
The key design decision is selecting a metric to measure how well the robot trajectory mimics the human demonstration. Importantly, we need to consider both the human-robot embodiment gap and the complex interaction between the robot hand and the object, a key difference from prior works that perform pure kinematic retargeting without considering the object. 

Inspired by OmniRetarget~\citep{yang2025omniretarget}, we define an interaction mesh-based retargeting objective to encourage robot motion to preserve the semantics of manipulating the object, such as the spatial and contact relationships between the robot hand and the object.
We define semantic keypoints on the human hand and the object as $\mathcal{P}_t^h$ and $\mathcal{P}_t^o$, respectively. They constitute the set of source points in the demonstration $\tilde{\mathcal{P}}_t=\mathcal{P}_t^o\cup \mathcal{P}_t^h$. Given the robot configuration $\mathbf{q}_t$, we define a set of robot keypoints $\mathcal{P}_t^r(\mathbf{q}_t)$, which semantically correspond to the human hand keypoints $\mathcal{P}_t^h$. The robot keypoints $\mathcal{P}_t^r(\mathbf{q}_t)$ and the original object keypoints $\mathcal{P}_t^o$ constitute the set of target points $\mathcal{P}_t(\mathbf{q}_t)=\mathcal{P}_t^o\cup \mathcal{P}_t^r(\mathbf{q}_t)$. We measure the interaction-preserving property by the deformation of interaction meshes (see Fig.~\ref{fig:interaction_mesh} for an illustration). The optimization program is formulated as  
\begin{equation}
    \label{eq:retargeting}
\begin{aligned}
    \bar{\mathbf{q}}_{0:T-1} = \mathop{\arg\min}_{\mathbf{q}_{0:T-1}}&\sum_{t=0}^{T-1} D\left( \mathcal{M}(\tilde{\mathcal{P}}_t), \mathcal{M}(\mathcal{P}_t(\mathbf{q}_t))\right) + \lambda\sum_{t=1}^{T-1} \left\| \mathbf{q}_t - \mathbf{q}_{t-1} \right\|_2^2 \\
    \text{s.t. } & \quad \mathbf{q}_{0:T-1} \in \mathcal{Q},
\end{aligned}
\end{equation}
where $\lambda$ is a weighting factor and $\mathcal{Q}$ is the feasible set of the robot configuration, including the kinematic constraints of the robot and collision avoidance constraints.
The connectivity between the keypoints $\mathcal{P}$ are decided by Delaunay tetrahedralization~\citep{si2005meshing}
to form the interaction mesh $\mathcal{M}(\mathcal{P})$ with vertices $\mathcal{P}$ and edges $\mathcal{E}$. We minimize the deformation energy between the two meshes, which is defined as the change of the Laplacian coordinates of the meshes: 
\begin{equation}
    D\left( \mathcal{M}(\tilde{\mathcal{P}}_t), \mathcal{M}(\mathcal{P}_t(\mathbf{q}_t))\right) = \sum_{i} \left\| L_i(\tilde{\mathcal{P}}_t) - L_i(\mathcal{P}_t(\mathbf{q}_t)) \right\|_2,
\end{equation}
where $L_i(\mathcal{P})$ is the Laplacian coordinate of the $i$-th vertex in the mesh $\mathcal{M}(\mathcal{P})=(\mathcal{P}, \mathcal{E})$, defined as
\begin{equation}
    L_i(\mathcal{P}) = \mathbf{p}_i - \frac{1}{|\mathcal{N}_i|} \sum_{j\in\mathcal{N}_i} \mathbf{p}_j,
\end{equation}
where $\mathbf{p}_i$ is the coordinate of the $i$-th vertex and $\mathcal{N}_i=\{j\in\mathcal{P}: (i,j)\in\mathcal{E}\}$ is the neighborhood of the $i$-th vertex. The second term in the objective (Eq.~\eqref{eq:retargeting}) encourages temporal smoothness of the trajectory.
The program is solved sequentially for each timestep $t$ using a Sequential Quadratic Programming (SQP)-style solver. See App.~\ref{app:retargeting_details} for implementation details.

\subsection{Reference Motion-Guided Policy Learning}
\label{sec:policy_learning}
Using the retargeted robot trajectory $\bar{\mathbf{q}}_{0:T-1}$, i.e., the solution of Eq.~\ref{eq:retargeting},   
as the reference motion, we learn an RL policy to track it using a generalizable keypoint tracking reward and reference state initialization. We use $\bar{\cdot}$ to denote the ``nominal'' quantity from the retargeted trajectory to distinguish it from the actual states in RL.

\myparagraph{Action Space.} The policy $\pi_\theta$ learns a residual action on top of the reference motion. The control target is then computed as $
    \mathbf{q}_{t}^{\text{target}} = \bar{\mathbf{q}}_{t} + \boldsymbol{\alpha}\odot \pi_\theta(\bar{\mathbf{q}}_{t}, \mathbf{o}_t),
$
where $\bar{\mathbf{q}}_t$ is the nominal robot configuration from the retargeted trajectory, $\mathbf{o}_t$ is the observation, and $\boldsymbol{\alpha}$ is a scaling vector controlling the action magnitude for each degree of freedom, and $\odot$ denotes element-wise product. This control target is then sent to a low-level PD controller.

\myparagraph{Observation Space.} We use asymmetric actor-critic observations. The actor observation contains the configurations of the object and the robot, the last action $\mathbf{a}_{t-1}$, as well as a phase variable $\phi\in[0,1]$, with $\phi=0$ indicating the start of the motion and $\phi=1$ indicating the end of the motion~\citep{peng2018deepmimic}. The object configuration consists of 
the object pose, including position $\mathbf{p}_t^o\in\mathbb{R}^3$ and orientation $\mathbf{R}_t^o\in\mathbb{R}^6$, represented by Rot6D~\citep{zhou2019continuity}, and the joint position $\mathbf{q}_t^o\in\mathbb{R}^{n_o}$ for articulated objects, where $n_o$ is the number of object joints.
These values are obtained from a motion capture system. The robot proprioception consists of all joint positions $\mathbf{q}_{t-1:t}$ from the current and the previous timesteps. We use the observations that can be relatively reliably obtained in the real-world system and avoid using noisy sensor measurements such as velocity for the policy to reduce the sim-to-real transfer gap. In addition to the above-mentioned observations, the critic also has access to the fingertip positions and joint velocities of the robot.

\myparagraph{Reward Function.} Our reward function is designed to be generalizable to different tasks and objects, and to be easy to implement, without requiring an explicit contact prior as in~\citep{mandi2025dexmachina,de2025scaffolding}. Instead, the prior for hand-object interaction is implicitly encoded in the reference trajectory generated by the interaction-preserving retargeting process. The main reward is a proximity reward incentivizing close object tracking.
We leverage a keypoint-based distance metric to measure the discrepancy between the current object configuration and the target object configuration:
\begin{equation}
    \label{eq:object_tracking_error}
    \epsilon_{\text{object}} = \mathrm{dist}_{\text{KP}}(\mathcal{P}_t^{o},\bar{\mathcal{P}}_t^{o}) = \frac{1}{N_k}\sum_{i=1}^{N_k} \left\| \mathbf{p}_{t,i}^o - \bar{\mathbf{p}}_{t,i}^o \right\|_2,
\end{equation}
where $\mathbf{p}_{t,i}^o$ and $\bar{\mathbf{p}}_{t,i}^o$ are the current and target positions of the $i$-th keypoint on the object, and $\bar{\mathcal{P}}_t^{o} = \{\bar{\mathbf{p}}_{t,i}^o\}_i$.
Compared to translation and rotation errors defined for rigid bodies, this distance metric is generally applicable to both rigid and articulated objects.
The reward is then shaped with an exponential kernel:
$
    r_{\text{object}} = \exp\left(-\epsilon_{\text{object}}/\sigma\right),
$
where $\sigma$ controls the sharpness of the reward. We also include object velocity and wrist pose tracking rewards, as well as penalties for large action magnitude and action rate to incentivize smooth motion.
See App.~\ref{app:rl_details} for all RL training details.

\begin{wrapfigure}{r}{0.4\columnwidth}
    \centering
    \includegraphics[width=0.4\columnwidth]{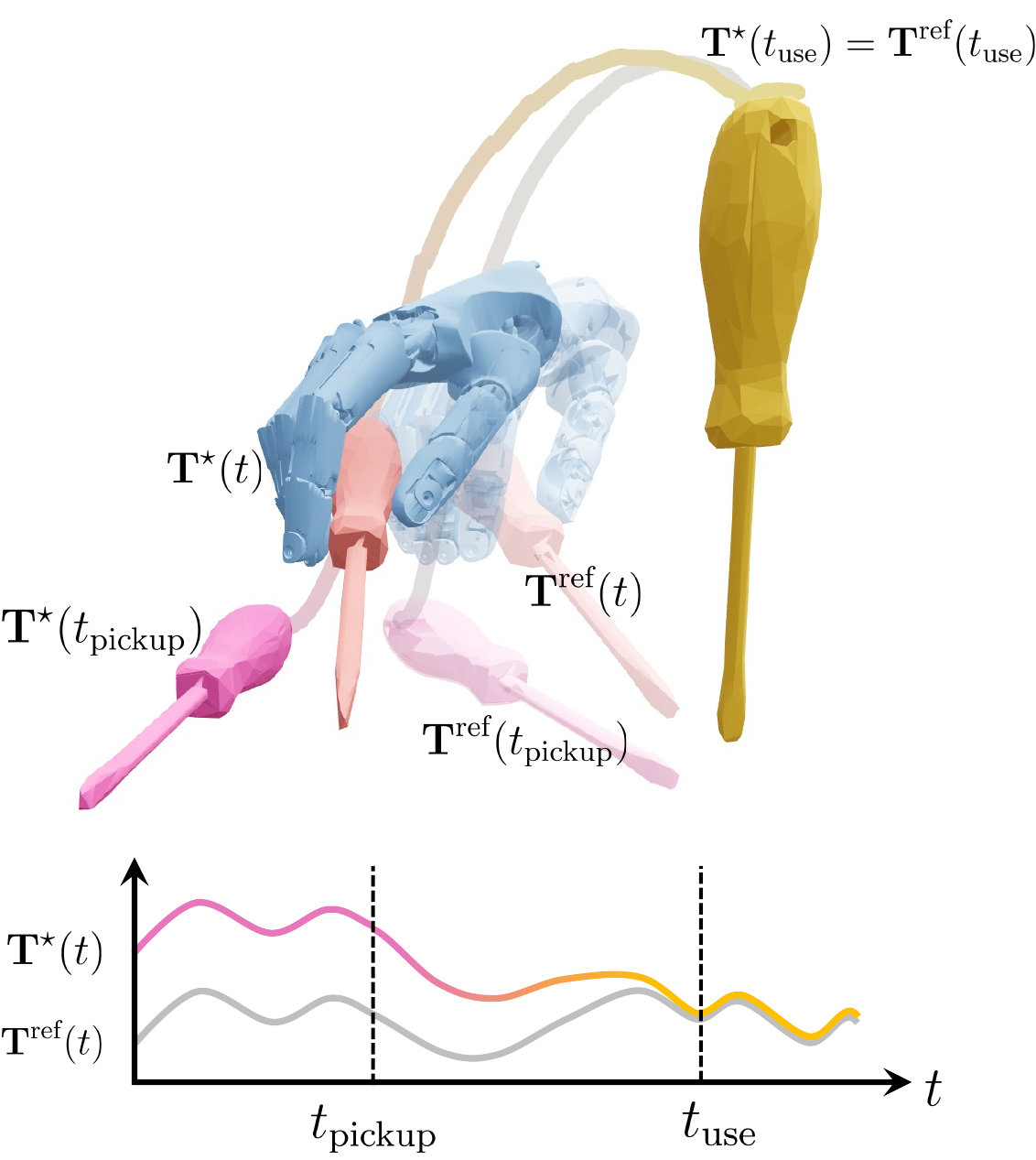}
    \vspace{-0.5em}
    \caption{\textbf{Data augmentation.} $\mathbf{T}^\text{ref}$ and $\mathbf{T}^\star$ denote the original and augmented retargeted trajectory, respectively. $\mathbf{T}^\star$ is interpolated towards $\mathbf{T}^\text{ref}$ between the timestep $t_\text{pickup}$ (when the object is picked up) and the timestep $t_\text{use}$ (when the subject starts to use the tool at a target position). See Appendix~\ref{app:data_augmentation} for more details.}
    \label{fig:augmentation}
    \vspace{-1em}
\end{wrapfigure}

\myparagraph{Reference Motion-Guided Training.} As motivated in Sec.~\ref{sec:problem_statement}, we adopt reference state initialization (RSI) \cite{peng2018deepmimic} and use the retargeted trajectory as the reference motion to guide the policy learning with a restart state distribution $\mu$. We sample the restart state from the retargeted trajectory, and initialize both the robot and the object to the corresponding states at the start of an episode, with a small perturbation to provide a broader coverage of potential high-value states. We also terminate the episode if the object deviates too much from the reference trajectory. Using a tight object deviation threshold is helpful to prevent the policy from deviating from the reference trajectory too much, since residual RL policies may not be able to recover from large deviations.

\myparagraph{Data Augmentation.}
Since our pipeline learns from a single human demonstration, the retargeted trajectory only covers one instantiation of the task, which limits the spatial generalization of the learned policy. We therefore perform data augmentation to synthesize a diverse set of reference trajectories for training a more robust and generalizable RL policy. We randomize the initial object configuration by perturbing its position and orientation around the demonstrated configuration ($\pm$5 cm for position and $\pm$30 degrees for orientation), which defines a rigid transformation $\Delta\mathbf{T}_0\in\mathrm{SE}(3)$ between the perturbed and the original initial object pose. To generate a new reference trajectory consistent with this perturbed start, we apply a time-varying rigid transformation $\Delta\mathbf{T}(t)$ to every state of the retargeted trajectory, obtained by interpolating between $\Delta\mathbf{T}_0$ at the start and the identity transformation at the goal. This warps the beginning of the trajectory toward the augmented initial state while smoothly returning it to the original goal state by the end.  Fig.~\ref{fig:augmentation} shows an example of the augmented trajectory for picking up and turning a screwdriver.

Different from OmniRetarget~\citep{yang2025omniretarget}, which reruns the retargeting procedure to generate a new robot trajectory for the augmented object trajectory, we directly apply the same rigid transformation to both the object pose and the robot hand root pose at each timestep without re-solving the optimization problem of retargeting. This enables us to dynamically generate an infinite number of augmented trajectories at RL training time while still preserving the relative spatial and contact relationships captured by the original retargeting process.

\myparagraph{Domain Randomization and Curriculum.} To further robustify the learned policy, we apply domain randomization to the physics simulation parameters (e.g., friction, motor gains, time delays, etc.) and add observation noise during RL training. In addition, we apply random pushes to both the object and the robot with a curriculum on the push strength. We also adopt gravity curriculum to gradually increase the gravity magnitude from zero to its full value, which is useful for tasks involving picking up objects. 

\secv
\section{Experiments}
\secv
We design experiments to answer the following key questions:
\begin{itemize}[itemsep=1pt, topsep=0pt, leftmargin=35pt, parsep=0pt]
    \item[\textbf{(Q1)}]\ Does interaction-preserving retargeting improve retargeting fidelity?
    \item[\textbf{(Q2)}]\ Does preserving interactions during retargeting improve downstream RL performance?
    \item[\textbf{(Q3)}]\
    Are the policies trained with \textsc{Regrind} able to transfer to real-world dynamics?
    \item[\textbf{(Q4)}]\ Can the policies replicate the demonstration from different initial states?
\end{itemize}

\subsection{Experiment Setup}
\myparagraph{Robots and Tasks.}
Our method is evaluated on two dexterous manipulation tasks: {\tt Scissors} from the ARCTIC dataset~\citep{fan2023arctic}, and {\tt Screwdriver} from our own data collection setup (see App.~\ref{app:demo_collection} for demonstration collection details), covering both rigid and articulated objects. Each task is evaluated on two anthropomorphic robot hands: a 16-DoF LEAP hand~\citep{shaw2023leap} and a 20-DoF WUJI hand. The robot hand is mounted on a 7-DoF UR5e robot arm. 
The tasks are illustrated in Fig.~\ref{fig:teaser}. In the scissors task, the robot picks up a pair of scissors and operates them at a desired orientation; in the screwdriver task, it grasps the screwdriver and turns it continuously while holding it upright. Due to its size, the LEAP hand cannot perform semantically similar grasps on normal-sized tools, so we use 3D-printed oversized objects for the LEAP hand, and real objects for the WUJI hand. 
To isolate the dynamics gap in sim-to-real transfer, we feed object poses from a motion capture system directly to the policy at deployment time, which minimizes perception errors and prevents them from acting as a confounder.

\myparagraph{Evaluation Metrics.} We use the following two metrics to evaluate the policy performance:
\begin{itemize}[itemsep=1pt, topsep=0pt, leftmargin=15pt, parsep=0pt]
    \item Object Tracking Error (Err.): The average distance between the current object keypoint positions and the target object keypoint positions (Eq.~\ref{eq:object_tracking_error}).
    \item Success rate (SR): The ratio of successful trials over the total number of trials. See App.~\ref{app:eval_details} for detailed success criteria for each task.
\end{itemize}

\myparagraph{Baselines.} We compare our method with two representative baselines for learning dexterous manipulation from human demonstrations.
\begin{itemize}[itemsep=1pt, topsep=0pt, leftmargin=15pt, parsep=0pt]
    \item SPIDER~\citep{pan2025spider}, a physics-based retargeting method that generates dynamically feasible robot trajectories. The retargeted trajectories can be executed as open-loop policies or used as reference motions for downstream RL.
    \item DexMachina~\citep{mandi2025dexmachina}, which also trains RL policies to track kinematic retargeted robot trajectories, but without considering the robot-object interaction semantics in the retargeting process.

\end{itemize}

In addition, we compare with a simple baseline that retargets the human motion to the robot with differential inverse kinematics (using Mink~\citep{zakka2026mink}) and then trains an RL policy with the same settings as our method, named Mink IK + RL.

\subsection{Results}
\myparagraph{(Q1) Retargeting Quality.}
\begin{figure}[htb]
    \centering
    \includegraphics[width=0.99\textwidth]{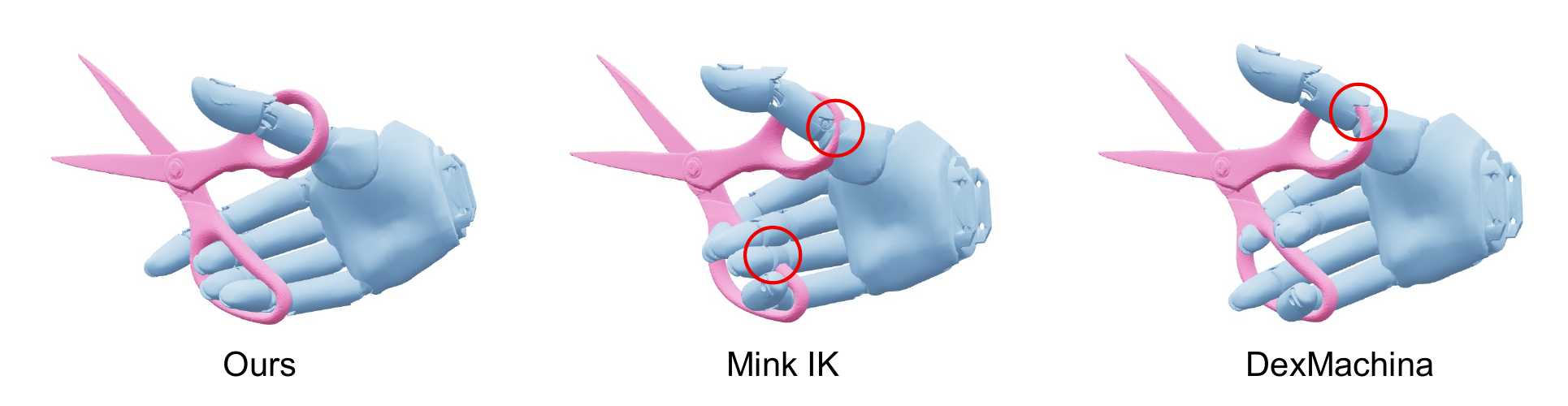}
    \caption{\textbf{Qualitative Retargeting Results.} Sample retargeted configurations produced by our method, Mink IK, and DexMachina for the {\tt WUJI-Scissors} task. Our method produces a human-like grasp that preserves the hand-object interaction semantics, while the other two baselines produce physically implausible or unstable grasps that are not suitable for RL training. The red circles highlight significant hand-object penetrations.}
    \label{fig:retargeting_results}
\end{figure}

We visualize the resulting robot configurations produced by different retargeting strategies for a qualitative comparison in Fig.~\ref{fig:retargeting_results}. Both the Mink IK + RL and DexMachina baselines use simple IK-based retargeting. They produce large robot-object penetrations in the IK solutions (highlighted in red circles), which can lead to instability when initializing the simulator to these states during RL training.
DexMachina uses a simulation-based post-processing step to eliminate these penetrations, but this projection can push the fingers to undesired positions, as no useful information is provided for the simulator to resolve the collision along the ``correct direction'' to preserver hand-object interaction. These suboptimal states are likely to harm RL training instead of providing consistent guidance. 

\myparagraph{(Q2) Simulation performance comparison.} 

\begin{table}[h]
    \centering
        \centering
        \caption{\textbf{Simulation Performance.} Object keypoint tracking error (Err.) and success rates (SR) on the 4 tasks in simulation. SPIDER results are averaged over 5 seeds; others are averaged over 3 seeds $\times$ 1024 rollouts.} 
        \label{tab:sim_comparison}
        \resizebox{\columnwidth}{!}{%
        \begin{tabular}{l|cr cr cr cr}
        \toprule 
        \multirow{2}{*}{\textbf{Method}}        &  \multicolumn{2}{c}{\tt LEAP-Scissors}  & \multicolumn{2}{c}{\tt LEAP-Screwdriver}  & \multicolumn{2}{c}{\tt WUJI-Scissors} & \multicolumn{2}{c}{\tt WUJI-Screwdriver}  \\
                                  \cmidrule(lr){2-3}              \cmidrule(lr){4-5}              \cmidrule(lr){6-7}              \cmidrule(lr){8-9}
                               &     {Err. (mm) $\downarrow$} & {SR $\uparrow$} & {Err. (mm) $\downarrow$} & {SR $\uparrow$} & {Err. (mm) $\downarrow$} & {SR $\uparrow$} & {Err. (mm) $\downarrow$} & {SR $\uparrow$} \\
        \midrule
            \textsc{Regrind} (Ours)& \textbf{\ms{5.6}{6.4}} & \textbf{\ms{99.8}{0.3}\%}  & \textbf{\ms{5.4}{5.8}} & \textbf{\ms{99.7}{0.0}\%}  & \textbf{\ms{5.3}{13.1}} & \textbf{\ms{98.7}{1.3}\%}  & \ms{6.5}{15.5} & \ms{98.8}{1.3}\%  \\
            SPIDER	           & \ms{176.2}{1.6}    & \ms{0.0}{0.0}\%  & \ms{134.5}{78.9}    & \ms{0.0}{0.0}\%  & \ms{119.0}{32.7}    & \ms{0.0}{0.0}\%  & \ms{116.9}{16.1}    & \ms{0.0}{0.0}\% 	\\
            DexMachina	       & \ms{10.1}{9.5} & \ms{22.3}{17.7}\% & \ms{8.3}{6.7} & \textbf{\ms{99.7}{0.1}\%} & \ms{67.2}{31.8} & \ms{0.0}{0.0}\% & \textbf{\ms{5.9}{9.7}} & \textbf{\ms{99.3}{0.1}\%} \\
            Mink IK + RL	   & \ms{12.6}{21.4} & \ms{2.0}{2.9}\% & \ms{17.5}{4.5} & \ms{0.0}{0.0}\% & \ms{38.9}{29.1} & \ms{0.0}{0.0}\% & \ms{10.0}{12.0} & \ms{3.1}{1.3}\% \\
        \bottomrule
        \end{tabular}}
\end{table}

Table~\ref{tab:sim_comparison} shows the performance of our method and the baselines on the four tasks in simulation. For all methods except SPIDER, we report the tracking error and success rate of the final RL policies. Across many hyperparameter settings, SPIDER produces object trajectories that deviate significantly from the original demonstration and are not useful initializations for residual RL. However, since SPIDER produces physically plausible, MPC-style trajectories that attempt to track the demonstration, we compare its retargeted trajectories against the trajectories produced by our closed-loop RL policies. Our method consistently achieves low object tracking error and near perfect success rates on all tasks. DexMachina also achieves near perfect success rates on the screwdriver tasks. However, its performance degrades on the more challenging scissors tasks, which involve complex object geometry and hand-object interactions. Both the Mink IK + RL and SPIDER baselines struggle with the contact-rich tasks considered here. 

\begin{table*}[htb]
    \centering
        \centering
        \caption{{\textbf{Real-World Performance.} Object keypoint tracking error (Err.) and success rates (SR) in the real world.}}
        \label{tab:real_performance}
        \vspace{0.5em}
        \resizebox{\columnwidth}{!}{%
        \begin{tabular}{l|cc cc cc cc}
        \toprule 
        \multirow{2}{*}{\textbf{Method}}        &  \multicolumn{2}{c}{\tt LEAP-Scissors}  & \multicolumn{2}{c}{\tt LEAP-Screwdriver}  & \multicolumn{2}{c}{\tt WUJI-Scissors} & \multicolumn{2}{c}{\tt WUJI-Screwdriver}  \\
                                  \cmidrule(lr){2-3}              \cmidrule(lr){4-5}
                                  \cmidrule(lr){6-7}              \cmidrule(lr){8-9}
                               &     {Err. (mm) $\downarrow$} & {SR $\uparrow$} & {Err. (mm) $\downarrow$} & {SR $\uparrow$} & {Err. (mm) $\downarrow$} & {SR $\uparrow$} & {Err. (mm) $\downarrow$} & {SR $\uparrow$} \\
        \midrule
            \textsc{Regrind} (Ours)	& \textbf{\ms{27.9}{2.0}}    & \textbf{9 / 10}  & \textbf{\ms{16.0}{2.9}}    & \textbf{10 / 10}  & \ms{65.2}{21.6}    & 0 / 10  & \textbf{\ms{9.6}{2.5}}    & \textbf{9 / 10}  \\
            DexMachina	       & \ms{228.7}{99.9}    & 0 / 10 & \ms{122.7}{85.4}    & 2 / 10 & ---    & --- & \ms{47.3}{45.0}    & 5 / 10 \\
            Mink IK + RL	   & ---    & ---  & ---    & ---  & ---    & ---  & \ms{41.8}{47.2}   & 0 / 10 	\\
        \bottomrule
        \end{tabular}}
\end{table*}

\myparagraph{(Q3) Sim-to-real transfer.} We deploy the RL policies that achieve reasonable performance in simulation to the real world and show the performance in Table~\ref{tab:real_performance}. Our method transfers reliably on three of the four tasks: \texttt{LEAP-Scissors}, \texttt{LEAP-Screwdriver}, and \texttt{WUJI-Screwdriver}. Surprisingly, despite strong simulation performance, the DexMachina policies transferred poorly on both screwdriver tasks. One hypothesis is that, without the stronger regularization provided by our interaction-preserving retargeting process, DexMachina policies are more prone to exploiting simulation artifacts, resulting in unstable grasps and overly aggressive motions that overshoot or hit the table during real-world deployment. Instead, our method produces human-like grasps that are robust to the real-world environment. None of the methods transferred successfully for the {\tt WUJI-Scissors} task, likely due to the larger sim-to-real gap with the non-backdrivable motors on the WUJI hand and the inaccurate mesh of the real scissors.

\myparagraph{(Q4) Generalization to Different Initial Configurations.}

\begin{wraptable}[10]{r}{0.5\textwidth}
    \centering
    \vspace{-6pt}
    \caption{\textbf{Generalization Performance.} Object keypoint tracking error (Err.) and success rates (SR) of our method on randomized initial configurations in the real world. 
    }
    \vspace{0.5em}
    \label{tab:generalization_performance}

    \resizebox{\linewidth}{!}{%
    \begin{tabular}{l|cr}
    \toprule 
    {\textbf{Task}}   &  {Err. (mm) $\downarrow$} & {SR $\uparrow$}  \\
    \midrule
        {\tt LEAP-Scissors}      & \ms{27.4}{2.7}    & 8 / 10  \\
        {\tt LEAP-Screwdriver}   & \ms{16.3}{2.1}    & 10 / 10  \\
        {\tt WUJI-Screwdriver}   & \ms{10.3}{2.2}    & 9 / 10   \\
    \bottomrule
    \end{tabular}%
    }
\end{wraptable}
To test the generalization capability of our policies beyond the demonstrated configurations, we evaluate the policies with randomly sampled initial object configurations. The position perturbations are sampled uniformly within $\pm$5 cm along the x and y axes, and the orientation perturbations are sampled uniformly within $\pm$30 degrees around the z axis. Table~\ref{tab:generalization_performance} shows the performance of our method on three tasks in the real world. We observe that the policies trained with our method generalize well to different initial states, achieving similar performance to the performance on the demonstrated configurations. This demonstrates the effectiveness of our simple dynamic data augmentation strategy at RL training to improve the robustness and generalizability of the policy. 

\secv
\section{Limitations}
\secv
\label{sec:limitations}

We are optimistic about retargeting-based learning for dexterous manipulation, as we have demonstrated promising results on challenging tasks and gained a better understanding of the challenges of sim-to-real transfer in contact-rich manipulation. At the same time, several limitations remain. First, our method relies on motion capture to obtain the object state at deployment time. Distilling the state-based RL policy into a vision-based policy is a natural next step to make the system more applicable to in-the-wild scenarios.
Second, our pipeline still requires careful system identification to enable sim-to-real transfer; a promising direction is to adapt the policy at test time using context inferred from interactions with the real world.

\acknowledgments{
 This project was supported in part by the Department of the Navy, Office of Naval Research under ONR award number N00014-25-1-2086.
 Model training was performed using the Unicorn shared computing cluster at Cornell University. Yunhai Feng would like to thank the PoRTaL Lab for support with motion capture prototyping, Paige Yun for help with the motion capture infrastructure, and Samuel Jin and other members at the Praxis Lab for helpful technical discussion.}

\clearpage

\bibliography{example}  

\clearpage

\appendix
\section{Motion Retargeting Details}
\label{app:retargeting_details}

\subsection{Keypoints and Correspondences}
\begin{figure}[ht]
    \centering
    \includegraphics[width=0.99\textwidth]{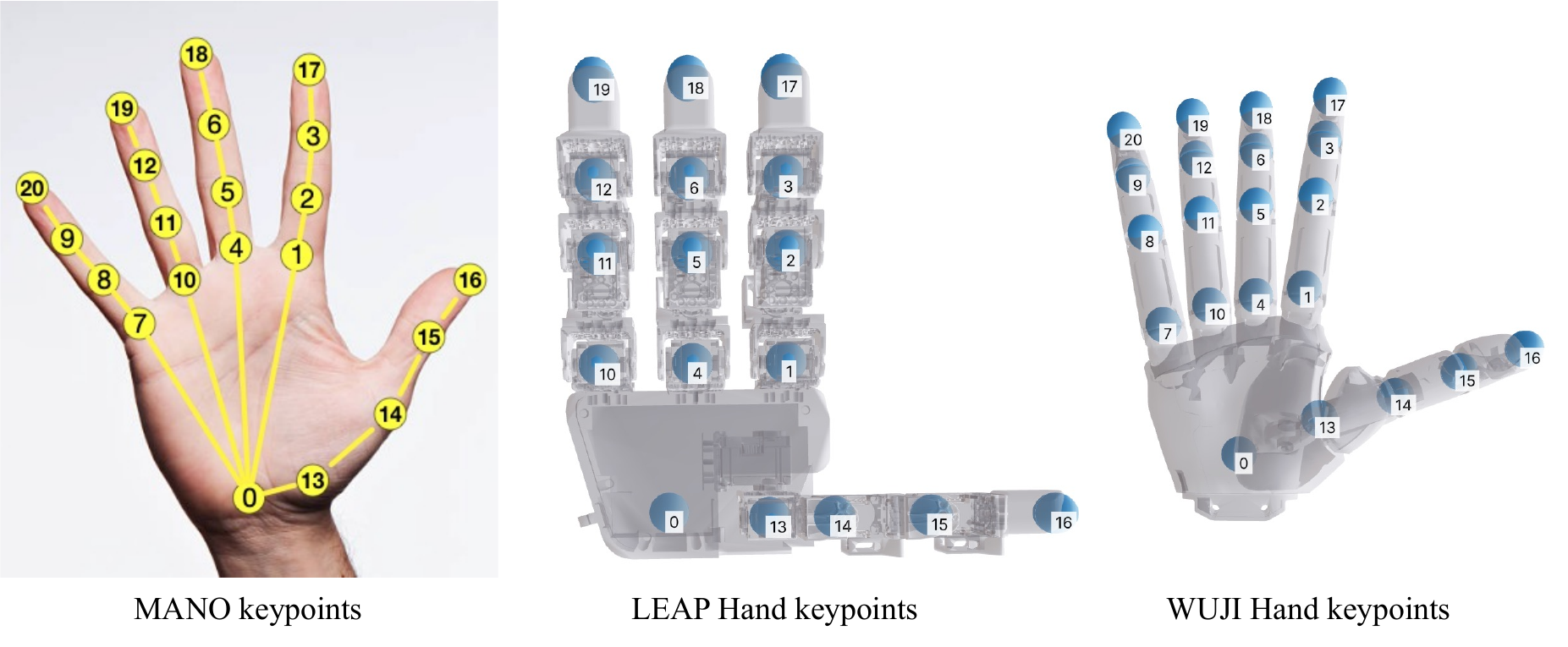}
    \caption{\textbf{Hand keypoint correspondence.} MANO human hand keypoints and the corresponding robot keypoints on the LEAP hand and WUJI hand.}
    \label{fig:keypoint_correspondence}
\end{figure}

To construct the interaction meshes, we sample 50 keypoints on each object. For scissors, we sample in the region of the handles where the fingers are making contact with. For screwdriver, we uniformly sample on the entire object surface. We define 21 hand keypoints on the joints, following the widely used MANO model\footnote{\url{https://mano.is.tue.mpg.de}}. Figure~\ref{fig:keypoint_correspondence} shows the MANO keypoints and the corresponding robot keypoints we define on the robot URDFs.

\subsection{Solver Details}
We follow OmniRetarget~\citep{yang2025omniretarget} for the interaction mesh-based retargeting process.
As in Sec.~\ref{sec:motion_retargeting}, $\mathbf{q}_t$ denotes the robot actuated configuration---the floating base together with all finger joint coordinates---while the demonstration object pose is held fixed and not optimized.
The trajectory-level program in Eq.~\eqref{eq:retargeting} is solved \emph{sequentially}: at each timestep $t$ we solve for $\mathbf{q}_t^*$, warm-started from $\mathbf{q}_{t-1}^*$.
Within each frame we run a small number of Sequential Quadratic Programming (SQP) iterations: the objective is quadratically approximated, the hard constraints in $\mathcal{Q}$ are linearized at the current iterate, and the convex subproblem is solved with MOSEK via Drake's \texttt{MathematicalProgram}\footnote{\url{https://github.com/RobotLocomotion/drake}}.

\subsection{Feasible Set $\mathcal{Q}$}

The kinematic constraints in Eq.~\eqref{eq:retargeting} are enforced as hard constraints on $\mathbf{q}_t$ at each timestep.
We write the feasible set as
\begin{equation}
    \mathcal{Q} =
    \bigl\{
        \mathbf{q}_t :
        \mathbf{q}_{\min} \le \mathbf{q}_t \le \mathbf{q}_{\max},\;
        \mathbf{v}_{\min}\,\Delta t \le \mathbf{q}_t - \mathbf{q}_{t-1} \le \mathbf{v}_{\max}\,\Delta t,\;
        \phi_j(\mathbf{q}_t) \ge 0,\;\forall j
    \bigr\},
    \label{eq:feasible-set}
\end{equation}
where $\mathbf{q}_{\min}$ and $\mathbf{q}_{\max}$ are joint limits, $\mathbf{v}_{\min}$ and $\mathbf{v}_{\max}$ bound the per-frame configuration change, $\Delta t$ is the demonstration timestep, and $\phi_j(\mathbf{q}_t)$ is the signed distance for the $j$-th collision pair.

\paragraph{Joint limits.}
We require $\mathbf{q}_{\min} \le \mathbf{q}_t \le \mathbf{q}_{\max}$, with limits taken from the robot model.
For the floating base, quaternion components are constrained to $[-1,1]$ before renormalization.

\paragraph{Velocity bounds.}
To encourage smooth retargeted motion, we bound the frame-to-frame change as $\mathbf{v}_{\min}\,\Delta t \le \mathbf{q}_t - \mathbf{q}_{t-1} \le \mathbf{v}_{\max}\,\Delta t$, matching the velocity-limit constraint used in OmniRetarget.

\paragraph{Non-penetration.}
For each active collision pair $j$, we enforce non-penetration via $\phi_j(\mathbf{q}_t) \ge 0$.
In our dexterous manipulation setting we activate robot--object and robot--environment (table/ground) pairs.
Adjacent geometries that should not generate constraints (e.g.\ palm--finger links) are filtered at plant setup.

\section{RL Training Details}
\label{app:rl_details}

\subsection{Simulation}
We use IsaacSim 5.1.0\footnote{\url{https://github.com/isaac-sim/IsaacSim/releases/tag/v5.1.0}} for large-scale parallel simulation. Simulation uses $\Delta t_{\mathrm{sim}} = 1/120\,\mathrm{s}$ with control decimation $4$, so the policy is evaluated every $\Delta t = 4\Delta t_{\mathrm{sim}} = 1/30\,\mathrm{s}$. Episodes last $9.0\,\mathrm{s}$ ($\approx 270$ control steps).

\subsection{Reward Functions}
\label{app:reward_functions}

We train with a dense reward that compares the simulated state to time-indexed targets from a retargeted reference motion. At each control step $t$, the scalar reward is a weighted sum of shaped terms:
\begin{equation}
    r_t \;=\; \sum_{k} w_k \, \phi_k\!\bigl(\mathbf{e}_{k,t}\bigr),
    \label{eq:reward-sum}
\end{equation}
where $w_k$ is the term weight, $\mathbf{e}_{k,t}$ is the corresponding tracking or regularization error, and $\phi_k$ maps the error to $[0,1]$ (or a binary penalty). Most tracking terms use an exponential kernel on a scalar error $e$:
\begin{equation}
    \phi_{\mathrm{lin}}(e;\,\sigma) = \exp\!\left(-\frac{e}{\sigma}\right).
    \label{eq:shaping-lin}
\end{equation}
Object linear/angular velocity matching, action magnitude, and action-rate terms use a squared-error cost $c$ with
\begin{equation}
    \phi_{\mathrm{sq}}(c;\,\sigma) = \exp\!\left(-\frac{c}{\sigma^2}\right).
    \label{eq:shaping-sq}
\end{equation}

\begin{table}[ht]
\centering
\caption{Reward terms and hyperparameters.}
\label{tab:reward-hparams}
\small
\begin{tabular}{@{}lllll@{}}
\toprule
\textbf{Term} & \textbf{Error / cost} & \textbf{Shaping} & $\boldsymbol{\sigma}$ & \textbf{Weight $w_k$} \\
\midrule
Object keypoint position  & mean $\|\mathbf{p}_i-\bar{\mathbf{p}}_i\|_2$ over keypoints
    & \eqref{eq:shaping-lin} & 0.02 & 1.5 \\
Object linear velocity    & $\|\mathbf{v}-\bar{\mathbf{v}}\|_2^2$
    & \eqref{eq:shaping-sq} & 1.0 & 1.0 \\
Object angular velocity & $\|\boldsymbol{\omega}-\bar{\boldsymbol{\omega}}\|_2^2$
    & \eqref{eq:shaping-sq} & 3.14 & 1.0 \\
Wrist position            & $\|\mathbf{p}_{\mathrm{wrist}}-\bar{\mathbf{p}}_{\mathrm{wrist}}\|_2$
    & \eqref{eq:shaping-lin} & 0.02 & 0.05 \\
Wrist orientation         & quaternion error magnitude
    & \eqref{eq:shaping-lin} & 0.2 & 0.05 \\
Action $\ell_2$ magnitude & $\mathrm{mean}_j(a_j^2)$
    & \eqref{eq:shaping-sq} & 1.0 & 0.5 (WUJI only) \\
Action rate $\ell_2$      & $\mathrm{mean}_j(a_j-a_{j,t-1})^2$
    & \eqref{eq:shaping-sq} & 0.5 & 1.0 \\
Action out of bounds      & sum of violations outside $[-1,1]$
    & \eqref{eq:shaping-lin} & 1.0 & 1.0 \\
Early termination         & $\mathbb{I}[\text{terminated} \land \neg\text{demo end}]$
    & -- & -- & -10.0 \\
\bottomrule
\end{tabular}
\vspace{0.25em}
\raggedright
\end{table}

\subsection{Data Augmentation}
\label{app:data_augmentation}
Referernce state initialization (RSI) samples a random demo phase at reset. Trajectory augmentation applies a per-episode perturbation, blended out between timesteps $t_{\text{pickup}}, t_{\text{use}}$ on object and hand targets.

Let $(\Delta\mathbf{p}, \Delta\psi)$ be a sampled perturbation transformation, with $\Delta\mathbf{p}_{xy} \in [-5,5]^2\,\mathrm{cm}$, $\Delta\mathbf{p}_z=0$, and $\Delta\psi \in [-30^\circ,30^\circ]$. The blend weight (linear interpolation of perturbation strength) is computed as:
\begin{equation}
w(t) =
\begin{cases}
1, & t < t_{\mathrm{pickup}}, \\
1 - \dfrac{t - t_{\mathrm{pickup}}}{t_{\mathrm{use}} - t_{\mathrm{pickup}}}, & t_{\mathrm{pickup}} \le t \le t_{\mathrm{use}}, \\
0, & t > t_{\mathrm{use}}.
\end{cases}
\end{equation}
The reference poses $(\mathbf{p}^{\mathrm{ref}}_t, \mathbf{R}^{\mathrm{ref}}_t)$ are then transformed to the augmented trajectory as:
\begin{align}
\mathbf{p}^\star(t) &= \mathbf{p}^{\mathrm{ref}}_t + w(t)\,\Delta\mathbf{p}, \\
\mathbf{R}^\star(t) &= \mathbf{R}_z\!\bigl(w(t)\,\Delta\psi\bigr)\,\mathbf{R}^{\mathrm{ref}}_t.
\end{align}
Hand wrist and fingertip positions use the same rigid map about the unperturbed object root $\mathbf{p}^{\mathrm{ref}}_{\mathrm{obj},t}$:
\begin{equation}
\mathbf{p}^\star_{\mathrm{hand}}(t) = \mathbf{R}_z\!\bigl(w(t)\,\Delta\psi\bigr)\bigl(\mathbf{p}^{\mathrm{ref}}_{\mathrm{hand},t} - \mathbf{p}^{\mathrm{ref}}_{\mathrm{obj},t}\bigr) + \mathbf{p}^{\mathrm{ref}}_{\mathrm{obj},t} + w(t)\,\Delta\mathbf{p};
\end{equation}
finger joint references are unchanged.

\begin{table*}[t]
    \centering
    \caption{Detailed domain randomization and curriculum terms.}
    \label{tab:dr-summary}
    \small
    \setlength{\tabcolsep}{4pt}
    \begin{tabular}{@{}p{3.5cm}p{5.5cm}p{5.5cm}@{}}
    \toprule
    \textbf{DR term} & \textbf{LEAP} & \textbf{WUJI} \\
    \midrule
    \multicolumn{3}{@{}l}{\textit{Physics}} \\
    Object CoM offset &
    scissors: $x,y,z = \pm2.5,\pm1,\pm0.5$\,cm;\newline
    screwdriver: $x,y,z=\pm0.5,\pm3,\pm0.5$\,cm &
    scissors: $x,y,z = \pm1.2,\pm0.5,\pm0.2$\,cm;\newline
    screwdriver: $x,y,z=\pm0.5,\pm2,\pm0.5$\,cm \\
    Object geometry scale &
    --- &
    scissors: $x,y\in[1.0,1.05]$, $z\in[0.95,1.0]$;\newline
    screwdriver: $x,y\in[1.0,1.05]$, $z{=}1.0$\\
    Robot/table/object friction &
    robot $[0.7,1.3]$; table, object $[0.5,1.5]$ &
    robot, table, object $[0.5,1.5]$ \\
    Robot/object mass scale &
    \multicolumn{2}{c}{$[0.95,1.05]$ / $[0.8,1.2]$} \\
    Joint $k_p,k_d$ scale &
    \multicolumn{2}{c}{$[0.9,1.1]$} \\
    Finger default-joint offset &
    $[-0.03,0.03]$\,rad &
    --- \\
    \midrule
    \multicolumn{3}{@{}l}{\textit{Curriculum}} \\
    Gravity &
    \multicolumn{2}{c}{$0 \to -9.81\,\mathrm{m/s^2}$ by env step $130\mathrm{k}$ (resampled each reset)} \\
    Random push &
    \multicolumn{2}{c}{inactive $<130\mathrm{k}$; max at $170\mathrm{k}$: $v\in[-0.5,0.5]$\,m/s, $\omega\in[-1,1]$\,rad/s; interval $[1,5]$\,s} \\
    \midrule
    \multicolumn{3}{@{}l}{\textit{Observations}} \\
    Additive noise &
    \multicolumn{2}{c}{pos $\pm 2$\,mm; rot $\pm 0.02$\,rad; joints $\pm 0.02$\,rad} \\
    Time lag &
    \multicolumn{2}{c}{$[0,2]$ control steps} \\
    \bottomrule
    \end{tabular}
\end{table*}

\subsection{Domain Randomization \& Curriculum}
\label{app:domain_randomization}
We apply domain randomization (DR) through Isaac Lab \emph{event terms}. Each parallel environment samples its own parameters. The detailed domain randomization setup is shown in Table~\ref{tab:dr-summary}.

\paragraph{Curriculum.} Gravity and random pushes depend on the number of environment steps. Gravity ramps from zero to $9.81\,\mathrm{m/s^2}$ by step $130\mathrm{k}$ (Table~\ref{tab:gravity-curriculum}). Pushes on robot and object activate from step $130\mathrm{k}$, reaching $\pm 0.5\,\mathrm{m/s}$ linear and $\pm 1.0\,\mathrm{rad/s}$ angular velocity at $170\mathrm{k}$, every $1$--$5\,\mathrm{s}$.

\begin{table}[h]
    \centering
    \caption{Gravity curriculum: sampled $g_z$ bounds (m/s$^2$; $g_x{=}g_y{=}0$). Columns are environment-step thresholds; each cell gives the resampled range at reset.}
    \label{tab:gravity-curriculum}
    \footnotesize
    \setlength{\tabcolsep}{3pt}
    \begin{tabular}{@{}l@{\hspace{2pt}}*{13}{c}@{}}
    \toprule
    Step $\geq$ & 0 & 20k & 30k & 40k & 50k & 60k & 70k & 80k & 90k & 100k & 110k & 120k & 130k \\
    \midrule
    $g_{\min}$ & 0 & 0 & 0.5 & 1.0 & 2.0 & 3.0 & 4.0 & 5.0 & 6.0 & 7.0 & 8.0 & 9.0 & 9.81 \\
    $g_{\max}$ & 0 & 1.0 & 2.0 & 3.0 & 4.0 & 5.0 & 6.0 & 7.0 & 8.0 & 9.0 & 9.81 & 9.81 & 9.81 \\
    \bottomrule
    \end{tabular}
\end{table}

\paragraph{Observation noise.} We apply uniform noise on policy observations at every timestep (position $\pm 2,\mathrm{mm}$, orientation $\pm 0.02,\mathrm{rad}$, joints $\pm 0.02,\mathrm{rad}$). We also apply continuous observation lag to model the delay in the real-world system (uniformly sampled in $[0,2]$ control steps per reset, with shared lag within object state, wrist pose, and hand joints).

\subsection{Termination Conditions}
The episode is early terminated if any of the following conditions is met: i) the object is out of the workspace; ii) the hand is moving too far away from the object; iii) the object is deviated from the target object configuration by more than a threshold (keypoint position error $> 0.15\,\mathrm{m}$); iv) for the WUJI hand, we also terminate the episode if the contact force between the fingertip and the table is greater than a threshold (10N) for safety concerns as the real WUJI hand motors are not back-drivable.

\subsection{RL Algorithm}
We use PPO implemented in RSL-RL\footnote{\url{https://github.com/leggedrobotics/rsl_rl}}. The hyperparameters are summarized in Table~\ref{tab:rl-algo-hparams}.

\begin{table}[ht]
    \centering
    \caption{RL algorithm and policy hyperparameters.}
    \label{tab:rl-algo-hparams}
    \small
    \begin{tabular}{@{}ll@{}}
    \toprule
    \textbf{Hyperparameter} & \textbf{Value} \\
    \midrule
    Steps per env per iteration & 24 \\
    Learning rate & $10^{-3}$ \\
    LR schedule & adaptive (target KL) \\
    Desired KL & 0.01 \\
    Learning epochs per iteration & 5 \\
    Mini-batches per iteration & 4 \\
    Discount $\gamma$ & 0.998 \\
    GAE $\lambda$ & 0.95 \\
    PPO clip $\epsilon$ & 0.2 \\
    Entropy coefficient & 0.002 \\
    Value loss coefficient & 1.0 \\
    Max gradient norm & 1.0 \\
    Actor MLP hidden dims & 1024, 512, 256, 128 \\
    Critic MLP hidden dims & 1024, 512, 256, 128 \\
    Activation & ELU \\
    Actor obs.\ normalization & yes \\
    Critic obs.\ normalization & yes \\
    Initial action noise std & 0.5 (scalar, diagonal Gaussian) \\
    \bottomrule
    \end{tabular}
    
    \vspace{0.35em}
    \raggedright
    \footnotesize
    With $N_{\mathrm{env}}=4096$ parallel environments, each iteration collects
    $N_{\mathrm{env}} \times 24 = 98{,}304$ transitions; each PPO update uses
    $5 \times 4 = 20$ optimizer passes with mini-batch size $98{,}304 / 4 = 24{,}576$. With the retargeted motion serving as the base action in residual RL, we initialize the last layer of the actor to produce zero-mean residual actions, and use a smaller initial action noise standard deviation (0.5).
    \end{table}

\section{Human Demonstration Collection}
\label{app:demo_collection}

\begin{figure}[ht]
    \centering
    \includegraphics[width=0.99\textwidth]{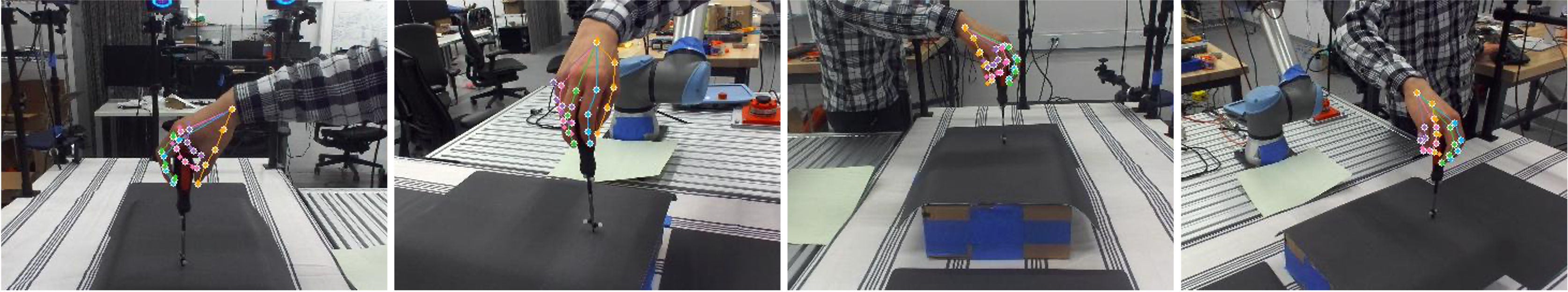}
    \caption{\textbf{Hand pose estimation.}}
    \label{fig:demo_collection}
\end{figure}

We use the data from the ARCTIC dataset\footnote{\url{https://github.com/zc-alexfan/arctic}} for the scissors task. For the screwdriver task, we collect the data from our own setup. We use a motion capture system to capture the object pose. To get the hand keypoint labels, we use the HaMeR hand pose estimation model\footnote{\url{https://github.com/geopavlakos/hamer}}, which produces the 21 MANO keypoints given an RGB image. We set up four calibrated cameras at different viewpoints, and triangulate the 3D hand keypoints from the 2D image keypoints for each timestep. See Fig.~\ref{fig:demo_collection} for an example of the estimated hand keypoints.

\section{Evaluation Details}
\label{app:eval_details}
\subsection{Task Success Criteria}
We define the task success criteria as follows.
\vspace{-0.2cm}
\paragraph{Scissors.}
Let $z_t$ be the object root height and $q_t$ the revolute joint angle.
Define the average blade direction $\mathbf{d}_t$ as the unit vector along the mean of the two blades' body $+x$ axes in world frame (the top blade rotated by $q_t$ about the joint axis).
An timestep is \emph{upright} if
\begin{align}
    z_t - z_0 &> 0.05 \cdot s \;\mathrm{m}\quad \text{(lifted off the table) }\\
    \mathbf{d}_t \cdot [0,-1,0]^\top &> 0.7 \quad \text{(properly oriented: body $x$ axis aligned with world $-y$)}
\end{align}
where $z_0$ is the initial height and $s$ is the task object scale ($s{=}2$ for the 3D-printed LEAP Hand scissors, $s{=}1$ for the real-size WUJI scissors).
Success holds if some contiguous upright segment contains both an opening and a closing sweep of at least $20^\circ$ each, i.e.\ maximum joint increase and decrease within that segment are each $\ge 20^\circ$.
\vspace{-0.2cm}
\paragraph{Screwdriver.}
Let $\mathbf{y}_t$ be the screwdriver body $+y$ axis in world frame.
A step is \emph{upright} when $-\mathbf{y}_t \cdot [0,0,1]^\top > 0.9$ (body $y$ axis aligned with world $-z$ axis).
While upright, we accumulate rotation about world $+z$ from successive orientations; the accumulator resets whenever the object is not upright.
Success requires the final timestep to be upright and the accumulated spin magnitude to be at least one full turn ($\ge 2\pi$).

\subsection{Baseline Details}
\paragraph{SPIDER} We use the official implementation of SPIDER\footnote{\url{https://github.com/facebookresearch/spider}} with the MuJoCo Warp workflow.
\vspace{-0.2cm}
\paragraph{DexMachina} We use a custom implementation of DexMachina-style retargeting and RL under the same framework as our method. We apply the same collision-aware post-processing step on top of the IK retargeted trajectories as in DexMachina. The downstream RL training is the same as our method.
\vspace{-0.2cm}
\paragraph{Mink IK + RL} This baseline simply replaces the retargeted trajectories in our method with the IK retargeted trajectories from Mink\footnote{\url{https://github.com/kevinzakka/mink}}.

\section{Real-World Robot System Details}
\begin{figure}[h]
    \centering
    \includegraphics[width=0.6\textwidth]{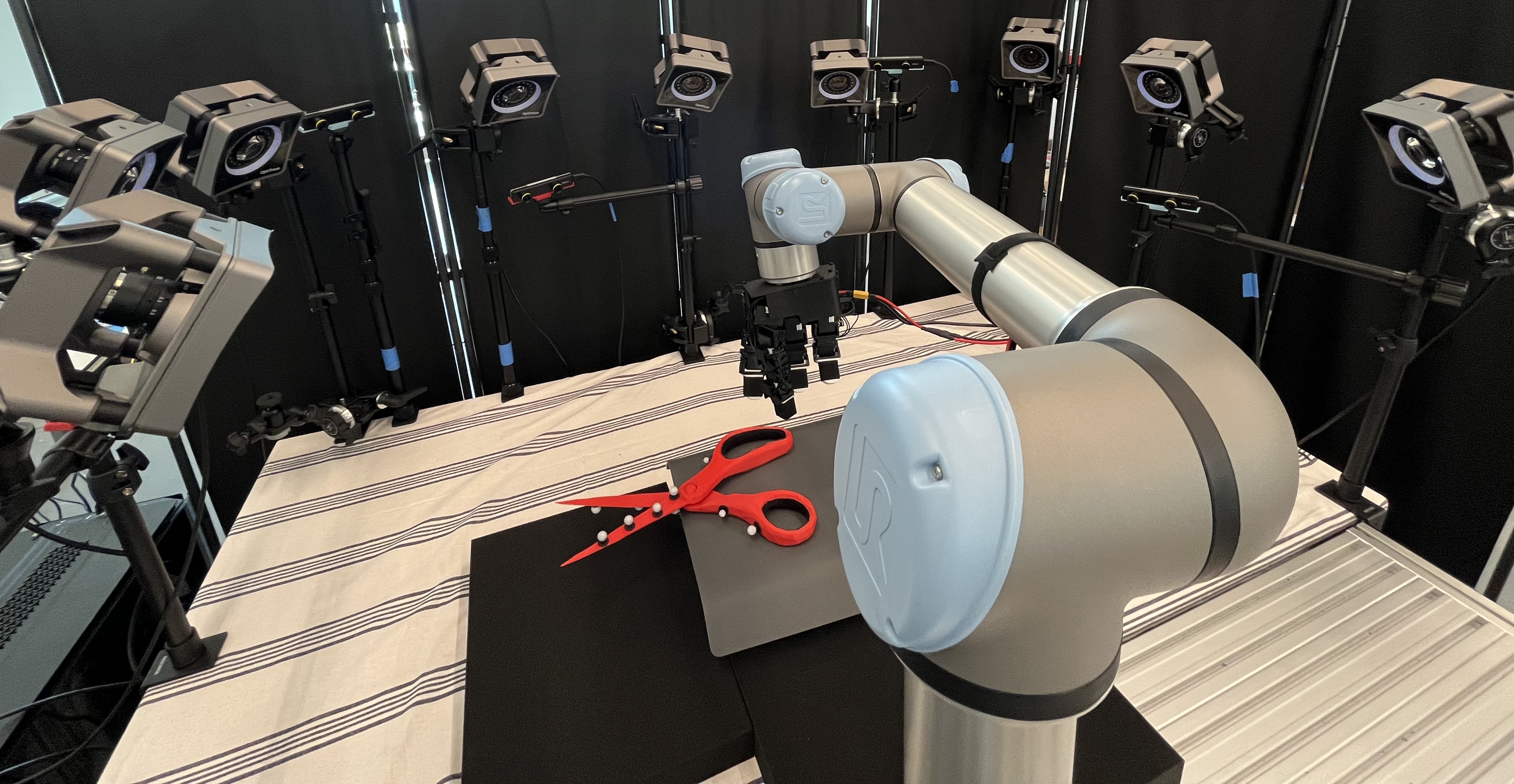}
    \caption{\textbf{Real-World Robot System.}}
    \label{fig:real_system}
\end{figure}

\vspace{-0.2cm}
\paragraph{Hardware Setup.} The system consists of a 7-DoF UR5e robot arm and a 16-DoF LEAP hand (or 20-DoF WUJI hand) and a motion capture system with 9 cameras for object pose estimation. See Fig.~\ref{fig:real_system} for the system setup.

\begin{figure}[h]
    \centering
    \includegraphics[width=0.99\textwidth]{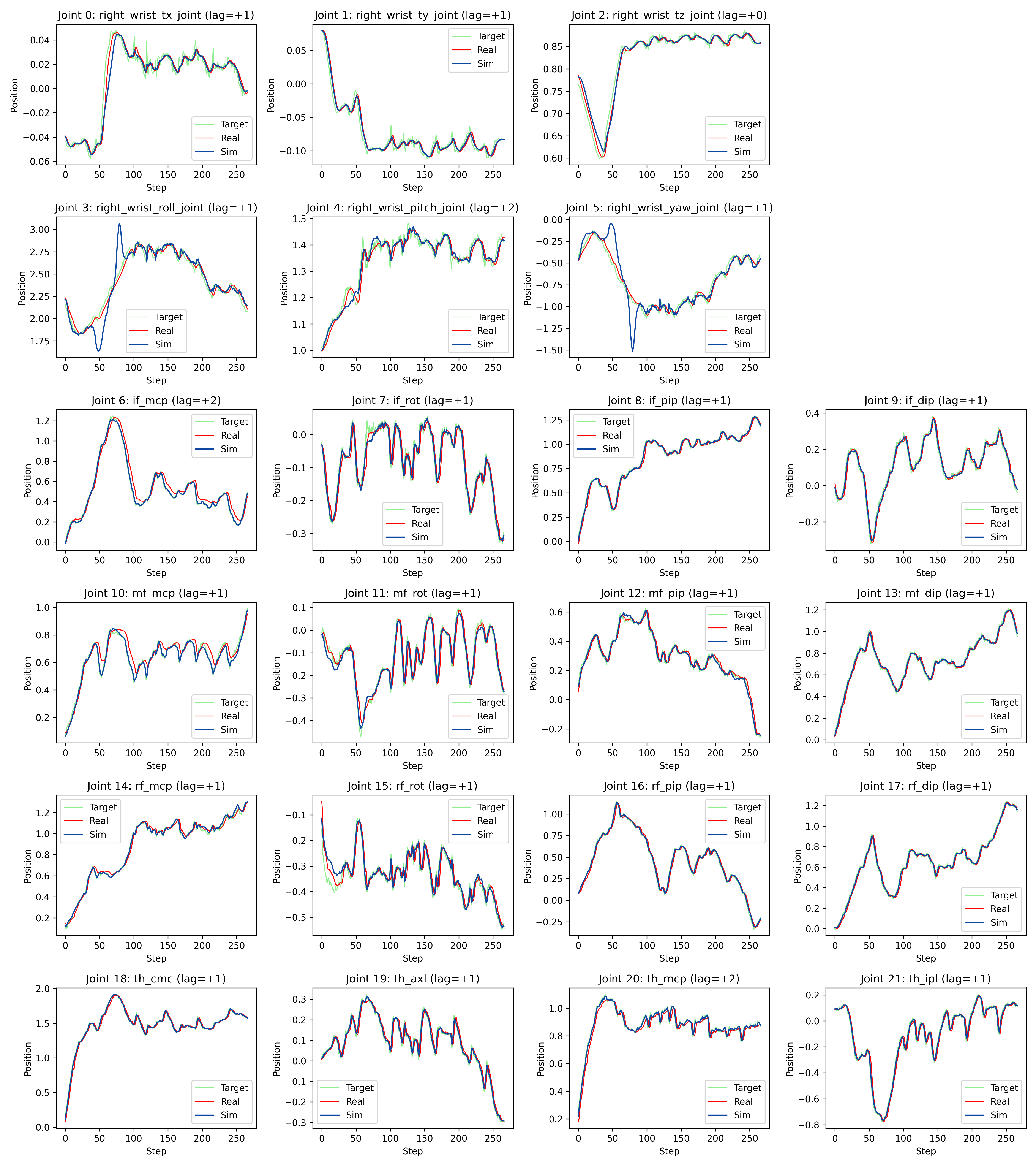}
    \caption{\textbf{Response curves for system identification.} Green: desired joint position; Red: real-robot response; Blue: simulated response.}
    \label{fig:system_identification}
\end{figure}

\paragraph{System Identification.} We perform system identification to align the robot dynamics between simulation and the real-world system. We record the policy rollouts in simulation, replay the joint targets on the real robot, and compare the response curves with the simulated ones. Fig.~\ref{fig:system_identification} shows the response curves for one trajectory on the LEAP hand. We observe that the real-robot response is slightly delayed by 1-2 timesteps ($\approx 30-60\,\mathrm{ms}$), which motivates us to apply a time lag in the observation pipeline in simulation to model the delay in the real-world system.

\end{document}